\documentclass[notitlepage,12pt,a4paper,onecolumn]{IEEEtran}

\usepackage{url}
\usepackage{graphicx}
\usepackage{color}
\usepackage{amsmath}
  \usepackage{amssymb}
 \DeclareMathOperator{\MMD}{MMD}
 \DeclareMathOperator{\RMMD}{RMMD}
 \DeclareMathOperator{\KFDA}{KFDA}
 \DeclareMathOperator{\cov}{cov}
 \DeclareMathOperator{\var}{var}
 \usepackage{url}
\makeatletter

\newcommand{\Rmnum}[1]{\expandafter\@slowromancap\romannumeral #1@}

\begin{document}
\title{Testing Hypotheses by Regularized Maximum Mean Discrepancy}
\author{
	{\bf Somayeh Danafar \IEEEauthorrefmark{1}\IEEEauthorrefmark{2}}, \and {\bf Paola M.V. Rancoita} \IEEEauthorrefmark{1}\IEEEauthorrefmark{3},\and {\bf Tobias Glasmachers} \IEEEauthorrefmark{4}\\ 
{\bf Kevin Whittingstall} \IEEEauthorrefmark{5} \IEEEauthorrefmark{6}, \and
{\bf J{\"u}rgen Schmidhuber} \IEEEauthorrefmark{1}\IEEEauthorrefmark{2}\\

	\IEEEauthorrefmark{1}IDSIA/SUPSI, Manno-Lugano, Switzerland\\
	\IEEEauthorrefmark{2}Universit\`{a} della Svizzera Italiana, Lugano, Switzerland\\
	\IEEEauthorrefmark{3}CUSSB, Vita-Salute San Raffaele University, Milan, Italy\\
	\IEEEauthorrefmark{4}Institut f{\"u}r Neuroinformatik, Ruhr-Universit{\"a}t Bochum, Germany\\
	\IEEEauthorrefmark{5}Dept. of Diagnostic Radiology, Universit\'{e} de Sherbrooke, QC, Canada\\
	 \IEEEauthorrefmark{6}Sherbrooke Molecular Imaging Center, Universit\'{e} de Sherbrooke, QC, Canada\\
}

\date{April 2013}
\maketitle

\begin{abstract}
 Do two data samples come from different distributions? Recent studies of this fundamental problem focused on embedding probability distributions into sufficiently rich characteristic Reproducing Kernel Hilbert Spaces (RKHSs), to compare distributions by the distance between their embeddings. We show that Regularized Maximum Mean Discrepancy (RMMD), our novel measure for kernel-based hypothesis testing, yields substantial improvements even when sample sizes are small, and excels at hypothesis tests involving multiple comparisons with power control. We derive asymptotic distributions under the null and alternative hypotheses, and assess power control. Outstanding results are obtained on: challenging EEG data, MNIST, the Berkley Covertype, and the Flare-Solar dataset.

\end{abstract}

\section{Introduction}

Homogeneity testing is an important problem in statistics and machine learning. It tests whether two samples are drawn from different distributions. This is relevant for many applications,  for instance, schema matching in databases \cite{Gretton07}, and speaker identification~\cite{Harchaoui}. Popular two-sample tests like  Kolmogorov-Smirnov ~\cite{Freidman} and Cramer-von-Mises~\cite{Rubner} are not capable of capturing statistical information of densities with high frequency features. Non-parametric kernel-based statistical tests such as Maximum Mean Discrepancy (MMD)~\cite{Gretton07,Gretton08} enable one to obtain greater power than such density based methods. MMD is applicable not only to Euclidean spaces $\mathbb{R}^n$, but also to groups and semigroups~\cite{Fukumizu09}, and to structures such as strings or graphs in bioinformatics, and robotics problems,  etc.~\cite{Borgwardt}. Here we consider a regularized version of MMD to address hypothesis testing.\\
 
With more than two distributions to be compared simultaneously, we face the multiple comparisons setting, for which statistical methods exist to deal with the issue of multiple test correction~\cite{Walsh}. Given a prescribed global significance threshold $\alpha$ (type~\Rmnum{1} error) for the set of all comparisons, however, the corresponding threshold per comparison becomes small, which greatly reduces the power of the test. In situations where one wants to retain the null hypothesis, tests with small $\alpha$ are not conservative.
Our main contribution is the definition of a regularized MMD (RMMD) method.\\

The regularization term in RMMD allows to control the power of the test statistic.
The regularizer is set {\bf provably optimal} for maximal power; there is no need for fine-tuning by the user.
RMMD improves on MMD through higher power, especially for small sample sizes, while preserving the advantages of MMD. {\bf Power control} enables us to look for true sets of null distributions among the significant ones in challenging multiple comparison tasks.\\

We provide experimental evidence of good performance on a challenging Electroencephalography (EEG) dataset, artificially generated periodic and Gaussian data, and the MNIST and Covertype datasets. We also assess power control with the Asymptotic Relative Efficiency (ARE) test.\\ 

The paper is organized as follows. In section~2, we elaborate on hypothesis testing and define maximum mean discrepancy (MMD) as a metric. We describe how to use MMD for homogeneity testing, and how to extend it to multiple comparisons. In section~3, we define RMMD for  hypothesis testing and compare it to MMD and Kernel Fisher Discriminant Analysis (KFDA), and assess power control through ARE. Additional empirical justification of our test on various datasets is presented in section~4.\\

\section{Statistical Hypothesis Testing}

A statistical hypothesis test is a method which, based on experimental data, aims to decide whether a hypothesis (called null or $H_0$) is true or false, against an alternative hypothesis ($H_1$). The level of significance $\alpha$ of the test represents the probability of rejecting $H_0$ under the assumption that $H_0$ is true (type~\Rmnum{1} error). A type~\Rmnum{2} error ($\beta$) occurs when we reject $H_1$ although it holds. \\

The {\bf power} of the statistical test is usually defined as $1-\beta$. A desirable property of a statistical test is that for a prescribed global significance level $\alpha$ the power equals one in the population limit. We divide the discussion of hypothesis testing into two topics: homogeneity testing and multiple comparisons.

\subsection{Maximum Mean Discrepancy (MMD)}

Embedding probability distributions into Reproducing Kernel Hilbert Spaces (RKHSs) yields a linear method that takes information of higher order statistics into account~\cite{Gretton07,Smola,Sriperumbudur08}. Characteristic kernels ~\cite{Fukumizu04,Sriperumbudur08,Fukumizu09} injectively map the probability distribution onto its mean element in the corresponding RKHSs. The distance between the \textbf{mean elements ($\mu$)} in the RKHS is known as MMD~\cite{Gretton07,Gretton08}. The definition of MMD~\cite{Gretton07} is given in the following theorem:\\

\textbf{Theorem 1.} \textit{Let ($\mathcal{X},\mathcal{B}$) be a metric space, and let $P$, $Q$ be two Borel probability measures defined on $ \mathcal{X}$. The kernel function $k:\mathcal{X}\times \mathcal{X}\rightarrow \mathbb{R}$ embeds the points $x\in \mathcal{X}$ into the corresponding reproducing kernel Hilbert space $\mathcal{H}$. Then $P = Q$ if and only if $\MMD(P,Q)=0$}, where
\begin{align}
\MMD (P,Q) &:=\|  \mu_{P}-\mu_{Q}\| _{\mathcal{H}}\nonumber\\
& =\|  E_{P}[k(x,.)]-E_{Q}[k(y,.)]\|  _{\mathcal{H}}\nonumber\\
&=(E_{x,x'\sim P}[k(x,x')]+E_{y,y'\sim Q}[k(y,y')]\nonumber\\
& -2E_{x\sim P,y\sim Q}[k(x,y)])^{\frac{1}{2}}.\
\end{align}

\subsection{Homogeneity Testing}

A two-sample test investigates whether two samples are generated by the same distribution. To do testing, MMD can be used to measure the distance between embedded probability distributions in RKHS. Besides calculating the distance measure, we need to check whether this distance is significantly different from zero. For this, the asymptotic distribution of this distance measure is used to obtain a threshold on MMD values, and to extract the statistically significant cases. We perform a hypothesis test with null hypothesis $H_{0}:P=Q$ and alternative $H_{1}:P\not=Q$ on samples drawn from two distributions $P$ and $Q$. If the result of MMD is close enough to zero, we accept $H_0$, which indicates that the distributions $P$ and $Q$ coincide; otherwise the alternative is assumed to hold. With $\alpha$ as a  threshold on the asymptotic distribution of the empirical MMD (when $P=Q$) , the ($1-\alpha$)-quantile of this distribution is statistically significant. Our MMD test determines it by means of a bootstrap procedure.

\subsection{Multiple Comparisons}

Statistical analysis of a data set typically needs testing many hypotheses. The multiple comparisons or multiple testing problem arises when we evaluate several statistical hypotheses simultaneously. Let $\alpha$ be the overall type~\Rmnum{1} error, and let $\bar{\alpha}$ denote the type~\Rmnum{1} error of a single comparison in the multiple testing scenario.  Maintaining the prescribed significance level of $\alpha$ in multiple comparisons yields $\bar{\alpha}$ to be more stringent than $\alpha$. Nevertheless, in many studies $\alpha= \bar{\alpha}$ is used without correction. Several statistical techniques have been developed to control $\alpha$ ~\cite{Walsh}. We use the Dunn-\^{S}id\'{a}k method: For $n$ independent comparisons in multiple testing, the significance level $\alpha$ is obtained by:  $\alpha=1-(1- \bar{\alpha})^{n}$. As $\alpha$ decreases, the probability of type~\Rmnum{2} error ($\beta$) increases and the power of the test decreases. This requires to control $\beta$ while correcting $\alpha$. To tackle this problem, and to control $\beta$, we define a new hypothesis test based on RMMD, which has higher power than the MMD-based test, in the next section. To compare the distributions in the multiple testing problem we use two approaches: one-vs-all and pairwise comparisons. In the one-vs-all case each distribution is compared to all other distributions in the family, thus $M$ distributions require $M-1$ comparisons. In the pairwise case each pair of distributions is compared at the cost of $\frac{M(M-1)}{2}$ comparisons.

\section{Regularized Maximum Mean Discrepancy (RMMD)}

The main contribution of this paper is a novel regularization of MMD measure called RMMD. This regularization aims to provide a test statistics with greater power (power closer to 1 with a prescribed type~\Rmnum{1} error~$\alpha$).
Erdogmus and Principe ~\cite{Erdogmus} showed that $-\log\|  \mu_P\| _\mathcal{H}^2$ is the Parzen window estimation of the Renyi entropy ~\cite{Renyi}. With RMMD we obtain a statistical test with greater power by penalizing the term $\| \mu_P\|  _\mathcal{H}^2+\|  \mu_Q\| _\mathcal{H}^2$. We formulate $\RMMD$ and its empirical estimator as follows:
 \begin{align}
 \RMMD (P,Q) &:= \MMD (P,Q)^2-\kappa_P \|   \mu_P \|  _{\mathcal{H}}^2-\kappa_Q \|   \mu_Q\|   _{\mathcal{H}}^2\\
 \widehat{\RMMD}(P,Q)&:= \|  \hat{\mu}_P-\hat{\mu}_Q \|   _\mathcal{H}^2-\kappa _P\|   \hat{\mu}_P \|  _{\mathcal{H}}^2-\kappa _Q\|   \hat{\mu}_Q \| _{\mathcal{H}}^2
 \end{align}
 
where $\kappa_P$, and $\kappa_Q$ are non-negative regularization constants. For simplicity we consider $\kappa_P=\kappa_Q=\kappa$ in many application, however, we can introduce prior knowledge about the complexity of distributions by choosing $\kappa_P \not= \kappa_Q$. The modified Jensen-Shanon divergence (JS) ~\cite{Fuglede} corresponding to RMMD is defined as:
\begin{align}
D(P,Q) &:= H_s(P,Q)- (\kappa +1)(H_s(P)+H_s(Q))
\end{align}
where $H_s$ denotes the (cross) entropy. Since $\kappa$ is positive, the absolute value of second term on the right-hand side of eq.~(4) increases, leading to a higher weight for the mutual information than for the entropy (vice versa if $\kappa$ would be lower than -1). \footnotemark
\footnotetext[1]{RMMD with negative-valued $\kappa$ can be used in clustering as a divergence to compare clusters. We achieve greater entropy with broader clusters. The resulting clustering method avoids overfitting with narrow clusters.}\\

Here we summarize the notation needed in the next section. Given samples $\{x_i\} _{i=1}^{n_1}$ and $\{y_i\} _{i=1}^{n_2}$ drawn from distributions $P$ and $Q$, respectively, the mean element, the cross-covariance operator and the covariance operator are defined as follows ~\cite{Fukumizu07,Gretton07}:
$\hat {\mu }_P=\frac{1}{n_1}\sum_{i=1}^{n_1}k(x_i,.)$, $\widehat{\Sigma}_{PQ}=\frac{n_1n_2}{n_1+n_2}(\hat{\mu}_P-\hat{\mu}_Q)\otimes(\hat{\mu}_P-\hat{\mu}_Q)$, and $\widehat{\Sigma} _P=\frac{1}{n_1} \sum_{i=1}^{n_1} (k(x_i,.) \otimes k(x_i,.))-(\hat{\mu}_P \otimes \hat{\mu}_P )$
, where $u \otimes v$ for $u,v\in \mathcal{H}$ is defined for all $f\in \mathcal{H}$ as $(u \otimes v)f=\langle v,f \rangle_{\mathcal{H}}u$. The quantities $\hat{\mu}_Q$ and $\widehat{\Sigma}_Q$ are defined analogously for the second sample $\{y_i\}_{i=1}^{n_2}$. The population counterparts, i.e., the population mean element and the population covariance operator are defined for any probability measure $P$ as $\langle \mu_P,f \rangle_{\mathcal{H}}=E[f(x)]$ for all $f\in \mathcal{H}$, and $\langle f,\Sigma_P g \rangle_{\mathcal{H}}=\cov _P [f(x),g(y)]$ for $f,g\in \mathcal{H}$.  From now on we call $\Sigma_B=\Sigma_{PQ}$ the \emph{between-distribution covariance}. The pooled covariance operator (which we call also the \emph{within-distribution covariance}) is denoted by: $\Sigma _W =\frac{n_1}{n_1+n_2}\Sigma _{P}+\frac{n_2}{n_1+n_2}\Sigma _{Q}$. 

\subsection{Limit Distribution of RMMD Under Null and Fixed Alternative Hypotheses}
 
Now we derive the distribution of the test statistics under the null hypothesis of homogeneity $H_0:P=Q$ (Theorem 2), which implies $\mu_P=\mu_Q$ and $\Sigma_{P}=\Sigma_{Q}=\Sigma_{W}$. Consistency of the test is guaranteed by the form of the distribution under $H_1:P\not=Q$ (Theorem 2). Assume that $\{x_i\}_{i=1}^{n_1}$ and $\{y_i\}_{i=1}^{n_2}$ are independent samples from P and Q, respectively (a priori they are not equally distributed). Let $z_i:=(x_i,y_i)$, $h(z_i,z_j):=k(x_i,x_j)+k(y_i,y_j)-k(x_i,y_j)-k(x_j,y_i)-h'(z_i,z_j)$, and $h'(z_i,z_j)=\kappa_Pk(x_i,x_j)+\kappa_Qk(y_i,y_j)$, and $\xrightarrow{D}$ denotes convergence in distribution. Without loss of generality we assume $n_1=n_2=n$, and $\kappa_P=\kappa_Q=\kappa$. The proofs hold even when $\kappa_P\neq\kappa_Q$. Based on Hoeffding~\cite{Hoeffding}, Theorem A (p. 192) and Theorem B (p. 193) by Serfling~\cite{Serfling}, we can prove the following theorem:\\

\textbf{Theorem 2.} \textit{If $E[h^2]<\infty$, under $H_1$, $\widehat{RMMD}$ is asymptotically normally distributed
 \begin{align}
  n^{\frac{1}{2}}(\widehat{RMMD}-RMMD)\xrightarrow{D}\mathcal{N}(0,\hat{\sigma}^2),\nonumber 
\end{align} 
with variance  $\hat{\sigma}^2=4(E_z[E_{z'}[ h(z,z')^2]]-E_{z,z'}^2[h(z,z')])$, uniformly at rate $1/\sqrt{n}$. Under $H_0$, the same convergence holds with $\hat{\sigma}^2=4~(E_z~[~E_{z'}~[~h'(z,z')^2~]~]-$ 
$E_{z,z'}^2[h'(z,z')])>0$.}\\

{\bf To increase the power} of our RMMD-based test we need {\bf to decrease the variance under $H_1$} in Theorem 2. The following Theorem can be used to obtain maximal power by setting $\kappa=1$. This will give us a fixed hyper-parameter---no need for user tuning. The optimal value of $\kappa$ decreases both the variance of $H_1$ and $H_0$ simultaneously and the fixed $\alpha$ is defined over the changed variance of $H_0$.\\

\textbf{Theorem 3.} \textit{The {\bf highest power} of RMMD is obtained for 
$\kappa = 1$.}\\

\textbf{Proof.} Let denote $A = k(x_i,x_j)+k(y_i,y_j)$ and $B = k(x_i,y_j)-k(x_j,y_i)$. Based on Theorem 2, the variance under $H_1$ is obtained by:
\begin{align}
\hat{\sigma}^2 &=4(E_z[E_{z'}[ h(z,z')^2]]-E_{z,z'}^2[h(z,z')])\nonumber\\
&=4(E[((1-\kappa )A-B)^2]-(E^2[(1-\kappa )A-B]))\nonumber\\
&=4((1-\kappa )^2 (E[A^2]-E^2[A])+ E[B^2]-E^2[B])\nonumber\\
&=4((1-\kappa )^2 \var(A) +\var(B)),
\end{align}
where $\var(A)$, and $\var(B)$ denote the variances. To get maximal power, we set
\begin{align}
\frac{\partial((1-\kappa^2)\var(A)+\var(B))}{\partial \kappa} =0, 
\end{align}
which yields $\kappa=1$.

\subsection{Comparison between RMMD, MMD, and KFDA}

According to Theorem~8 by Gretton et al.\ \cite{Gretton07}, under the null hypothesis the test statistics of MMD degenerates. This corresponds to $\hat{\sigma}^2=0$ in our Theorem~2. For large sample sizes the null distribution of MMD approaches in distribution as an infinite weighted sum of independent $\chi_1^2$ random variables, with weights equal to the eigenvalues of the within-distribution covariance operator $\Sigma_W$. If we denote the test statistics based on MMD by $\hat{T}_n^{\MMD}$, then $\hat{T}_n^{\MMD}\xrightarrow{D}C\sum_{l=1}^{\infty} \lambda_l(z_l^2-1)$, where $z_l\sim \mathcal{N}(0,2)$ are i.i.d.\ random variables, and $C$ is a scaling factor. 
Harchaoui et al.\ ~\cite{Harchaoui} introduced Kernel Fisher Discriminant Analysis (KFDA) as a homogeneity test by regularizing MMD with the within-distribution covariance operator. The maximum Fisher discriminant ratio defines this test statistic. The empirical KFDA test statistic is denoted as $\widehat{\KFDA}(P,Q)=\frac{n_1 n_2}{n_1+n_2}\|  \frac{\hat{\mu}_P-\hat{\mu}_Q}{(\hat{\Sigma}_W+\gamma_nI)^{\frac{1}{2}}}\|  _\mathcal{H}^2$. To analyze the asymptotic behaviour of this statistics under the null hypothesis, Harchaoui et al.\ \cite{Harchaoui} consider two situations regarding the regularization parameter $\gamma_n$: 1) one where $\gamma_n$ is held fixed, obtaining the limit distribution similar to MMD under $H_0$; 2) one where $\gamma_n$ tends to zero slower than $n^{-1/2}$. In the first situation the test statistic converges to 
$\hat{T}_n^{\KFDA(\gamma_n)}\xrightarrow{D}C\sum_{l=1}^{\infty}(\lambda_l+\gamma_n)^{-1}\lambda_l(z_l^2-1)$. Thus, the test statistics based on KFDA normalizes the weights of $\chi_1^2$ random variables by using the covariance operator as the regularizer. In comparison MMD is more sensitive to the information of higher order moments because of their bigger weights (larger eigenvalues of the covariance operator).
In the second situation (applicable in practice only for very large sample sizes) the test statistics converges to $\hat{T}_n^{\KFDA(\gamma_n)}\xrightarrow{D}\mathcal{N}(C,1)$, where $C$ is a constant.\\

The asymptotic convergence of the test statistic based on RMMD is $\hat{T}_n^{\RMMD} \xrightarrow{D}\mathcal{N}(0,\hat{\sigma}^2)$,
where $\hat{\sigma}^2$ is the variance of the function $h$ in Theorem 2. The precise analytical normal distribution obtains higher power in RMMD. Because of the divergence ($\sigma^2 = 0$ in the asymptotic distribution) for MMD and KFDA, they use an estimation of the distribution under the null hypothesis which looses the accuracy and affect the power. In contrast to MMD and KFDA, RMMD is consistent since the divergence under the null hypothesis does not happen any more. RMMD is the generalized form of the test statistics based on MMD, which we obtain for $\kappa=0$.  Moreover, by minimizing the variance of the normal distribution, we obtain the best power for $\kappa = 1$ and thus the hyper-parameter $\kappa$ is fixed without requiring tuning by the user.\\

In comparison to KFDA, RMMD does not require restrictive constraints to obtain high power. It also results in higher power than MMD and KFDA in cases with small sample size. The speed of power convergence in KFDA is $O_p(1)$, which is slower than $O_p(n^{-\frac{1}{2}})$ in RMMD when $n \rightarrow\infty$. \\

Regarding the computational complexity, for MMD a parametric model with lower order moments of the test statistics is used to estimate the value of MMD which degenerates under $H_0$, and which has no consistency or accuracy guarantee. In comparison, the bootstrap resampling and the eigen-spectrum of the gram matrix are more consistent estimates with computational cost of $O(n^2)$, where $n$ is the number of samples ~\cite{Gretton09}. For RMMD, the convergence of the test statistic to a Normal distribution enables a fast, consistent and straightforward estimation of the null distribution within $O(n^2)$ time without the need of using an estimation method.
The results of power comparison between these tests are reported in section~4.

\subsection{Asymptotic Relative Efficiency of Statistical Tests}

To assess the power control we use the asymptotic relative efficiency. This criterion shows that RMMD is a better test statistic and obtains higher power rather than KFDA and MMD with smaller sample size. Relative efficiency enables one to select the most effective statistical test quantitatively ~\cite{Nikitin}. Let $T$ and $V$ be test statistics to be compared. The necessary sample size for the test statistics T to achieve the power $1-\beta$ with the significance level $\alpha$ is denoted by $N_T(\alpha, 1-\beta)$. The relative efficiency of the statistical test $T$ with respect to the statistical test $V$ is given by:
 \begin{align}
 e_{T,V}(\alpha, 1-\beta) = N_V(\alpha , 1-\beta)/ N_T(\alpha, 1- \beta).
 \end{align}
 Since calculating $N_T(\alpha, 1-\beta)$ is hard even for the simplest test statistics, the limit value $e_{T,V}(\alpha; 1-\beta)$, as $1-\beta \rightarrow 1$, is used. The limiting value is called the Bahadur Asymptotic Relative Efficiency (ARE) denoted by $e^B_{T,V}$.
 \begin{align}
 e^B_{T,V}:=\lim_{~~~~~1-\beta \rightarrow 1}{e_{T,V}(\alpha, 1-\beta)},
 \end{align}
The test statistic $V$ is considered better than $T$, if $e_{T,V}$ is smaller than 1, because it means that $V$ needs  a lower sample size to obtain a power of $1-\beta$, for the given $\alpha$. In \cite{Harchaoui}, authors assessed the power control by means of analysis of local alternatives which work when we have very large sample size or when n tends to infinity. In this article, we focus our attention on the small sample size case, which is more challenging. In section 4, we compute $e^B_{\MMD,\RMMD}=\frac{N_{~\RMMD}}{N_{~\MMD}}$, $e^B_{\MMD,\KFDA}=\frac{N_{~\KFDA}}{N_{~\MMD}}$, and $e^B_{\KFDA,\RMMD}=\frac{N_{~\RMMD}}{N_{~\KFDA}}$ using artificial datasets and two types of kernels, and we obtain smaller ARE for RMMD rather than KFDA and MMD. This means RMMD gives higher power with much smaller sample size. Results for different data sets are reported in Table~2, Figure~2, and Figure~3.

\section{Experiments}

MMD ~\cite{Gretton07} was experimentally shown to outperform many traditional two-sample tests such as the generalized Wald-Wolfowitz test, the generalized Kolmogorov-Smirnov (KS) test ~\cite{Freidman}, the Hall-Tajvidi (Hall) test ~\cite{Hall}, and the Biau-Gy\"{o}rf test. It was shown ~\cite{Harchaoui} that KFDA outperforms the Hall-Tajvidi test. We select KS and Hall as traditional baseline methods, on top of which we compare RMMD, KFDA, and MMD. To experimentally evaluate the utility of the proposed hypothesis testing method, we present results on various artificial and real-world benchmark datasets.

\subsection{Artificial Benchmarks with Periodic and Gaussian Distributions}

Our proposed method can be used for testing the homogeneity of structured data, which is an advantage over traditional two-sample tests. We artificially generated distributions from Locally Compact Abelian Groups (periodic data) and applied our RMMD-test to decide whether the samples come from the same distributions or not. Suppose the first sample is drawn from a uniform distribution $P$ on the unit interval. The other sample is drawn from a perturbed uniform distribution $Q_\omega$ with density $1+\sin(\omega x)$. For higher perturbation frequencies $\omega$ it becomes harder to discriminate $Q_\omega$ from $P$. Since the distributions have a periodic nature, we use a characteristic kernel tailored to the periodic domain, $k(x,y)=\cosh(\pi-(x-y)_{\text{mod}~2\pi})$. For 200 samples from each distribution, the type~\Rmnum{2} error is computed by comparing the prediction to the ground truth over 1000 repetition. We average the results over 10 runs. The significance level is set to $\alpha=0.05$. We perform the same experiment with MMD, KFDA, KS and Hall. The powers of the homogeneity test for comparing $P$ and $Q_6$ with the above mentioned methods are reported in Table 1 as Periodic1. The best power is achieved by RMMD, and as expected, the results of kernel methods are better than traditional ones.\\

Since the selection of the kernel is a critical choice in kernel-based methods, we also investigated the usage of a different kernel and replaced the previous kernel with $k(x,y)=-\log(1-2\theta \cos(x-y)+\theta^2)$, where $\theta$ is a hyperparameter. We report the best results achieved by $\theta =0.9$ as Periodic2 in Table 1. The reader is referred to \cite{Danafar,Fukumizu09} for a detailed study on these kernels.\\

We also report the results on the toy problem of comparing two $25$-dimensional Gaussian distributions with $250$ samples, both with zero mean vector but with covariance matrix $1.5~I$ and $1.8~I$, respectively. This dataset is referred as Gaussian in Table~1.
 \begin{table*}[!ht]
\caption{The Power obtained on the periodic data, the Gaussian, the MNIST, Covertype, and Flare Solar datasets, by applying RMMD with $\kappa=0.8$ for the periodic data and $\kappa=1$ for the others, and KFDA with $\gamma=10^{-1}$.}
 \begin{center}
 \begin{normalsize}
 \begin{sc}
 \begin{tabular}{lcccccr}
 \hline
  & \bf{RMMD} & \bf{KFDA} & \bf{MMD} & \bf{KS} & \bf{Hall} \\
 \hline
Periodic1 & \bf{0.40$\pm$ 0.02} & 0.24$\pm$ 0.01 & 0.23$\pm$ 0.02 & 0.11$\pm$ 0.02 & 0.19 $\pm$ 0.04\\
Preiodic2 & \bf{0.83$\pm$ 0.03} & 0.66$\pm$ 0.05 & 0.56$\pm$ 0.05 & 0.11$\pm$ 0.02 & 0.19 $\pm$ 0.04\\
Gaussian & \bf{1.00}& 0.89 $\pm$ 0.03 & 0.88 $\pm$0.03 & 0.04 $\pm$ 0.02 & \bf{1.00} \\
MNIST & \bf{0.99$\pm$ 0.01} & 0.97$\pm$ 0.01 & 0.95$\pm$ 0.01 & 0.12 $\pm$ 0.04 & 0.77 $\pm$ 0.04\\
Covertype & \bf{1.00} & \bf{1.00} & \bf{1.00} & 0.98$\pm$0.02 & 0.00 \\
Flare-Solar & \bf{0.93} & 0.91 & 0.89 & 0.00 & 0.00 \\
\hline
\end{tabular}
\end{sc}
\end{normalsize}
\end{center}
\end{table*}

An investigation of the effect of kernel selection and tuning parameters ~\cite{Sriperumbudur09} showed that best results for MMD can be achieved by those kernels and parameters that obtain supreme value for MMD. Our reported results agree. The results of kernel-based test statistics (RMMD, KFDA, and MMD) are improved by kernel justification and parameter tuning, and in all cases RMMD outperform KFDA and MMD. For instance, the result of periodic kernel with tuned hyper-parameter $\theta$ is better than the one of the first periodic kernel without hyper-parameter (reported in Table 1 as Periodic2 and Periodic1, respectively). For Gaussian kernel-processed datasets, the median distance between data points provided the best results. We used the 5-fold cross validation procedure to tune the parameters in our experiment.\\

The effect of changing $\kappa$ on the power is simulated in two tests: first, by testing the similarity between the uniform distribution and $Q_4$, and second with $Q_6$. In both cases, the best power is obtained for $\kappa=0.8$. The results slightly differ from the theoretical value ($\kappa=1$) because of the relatively small sample sizes ($n_1=n_2=200$) used for the tests. For samples with larger sizes we obtained maximal power with $\kappa=1$. The results are depicted in Figure~1.

\begin{figure*}[!ht]
 \begin{center}
 \centerline{\includegraphics[scale=0.7]{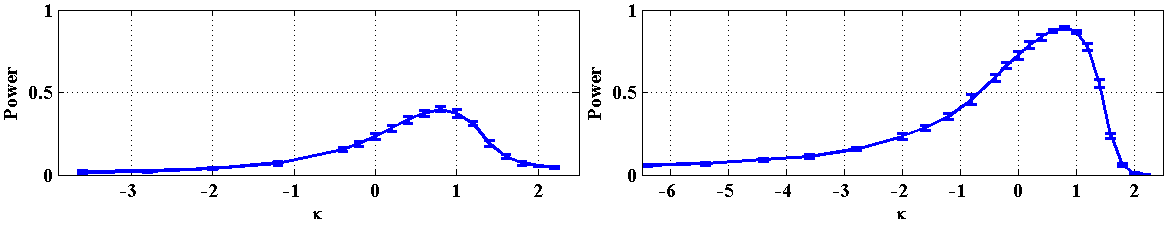}}
 \caption{ Effect of $\kappa$ on the power of the test. The alternatives are $Q_6$ in the left and $Q_4$ in the right figure.}
\label{fig2}
\end{center}
\end{figure*}
To assure that the statistical test is not aggressive for rejecting the null hypothesis, we reported the results of type \Rmnum{1} error for RMMD, KFDA, and MMD with different sample sizes in Figure 2. Both samples are supposed to be drawn from $Q_6$. We used Gaussian kernel with a variance equals to medium distance of data points. The results averaged over 100 runs and the confidence interval obtained by 10 replicates. RMMD obtains zero type \Rmnum{1} error with smaller sample sizes, and the results of KFDA and MMD are comparable.\\

\begin{figure*}[!ht]
 \begin{center}
 \centerline{\includegraphics[scale=0.5]{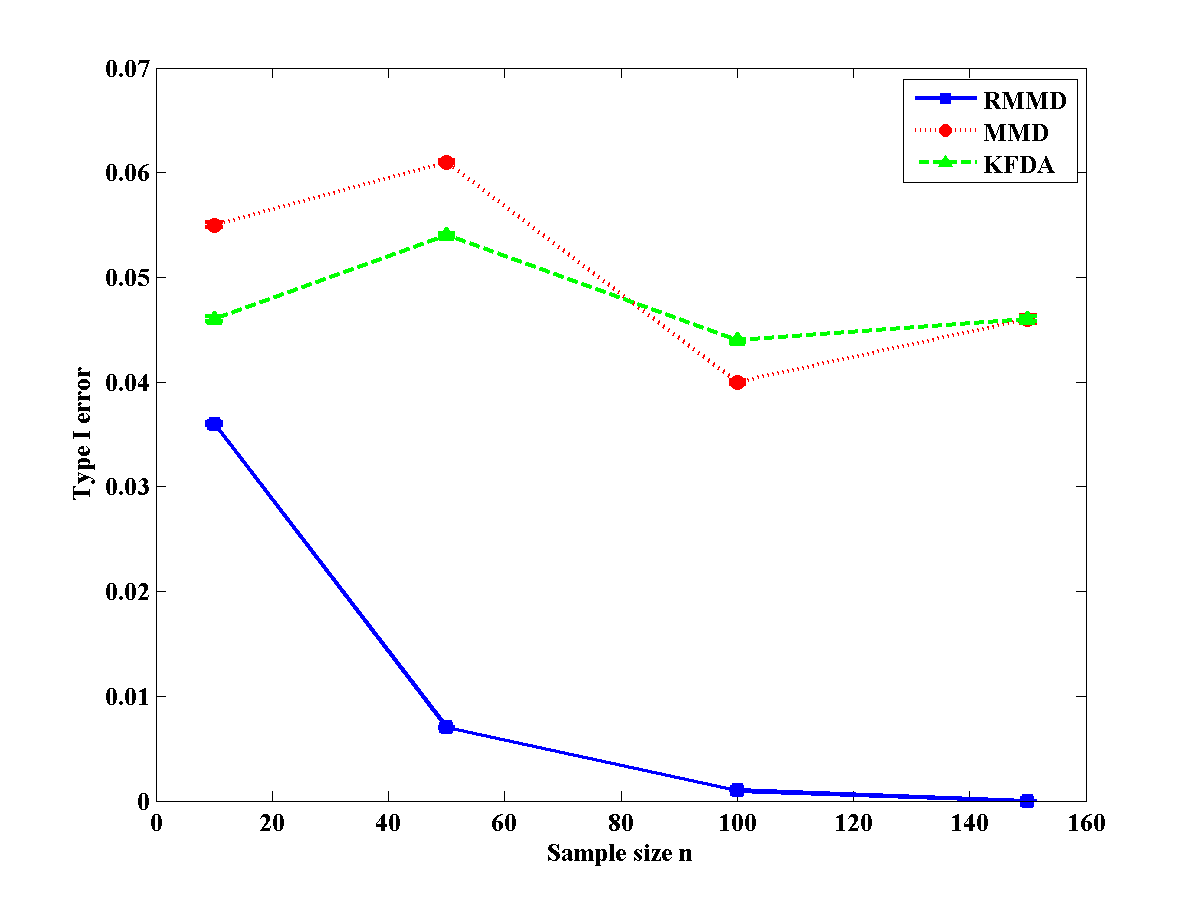}}
 \caption{Type \Rmnum{1} error changed based on different sample size \emph{n}.}
\label{fig2}
\end{center}
\end{figure*}
To assess the power control of the test statistics we also compared $e^B_{\MMD,\RMMD}$, $e^B_{\MMD,\KFDA}$, and $e^B_{\KFDA,\RMMD}$ under $H_1$ when $P$ is a uniform distribution and the alternative is $Q_6$. We obtained smaller ARE for RMMD rather than for KFDA and MMD. This means RMMD gives higher power with fewer samples. Table~2 shows the results, averaged over 1000 runs, for periodic data (Periodic1 and Periodic2). Figure 3 depicts the detailed results of the type~\Rmnum{2} error for RMMD, MMD, and KFDA based on different sample sizes~$n$. AREs are also calculated for more complex tasks. Consider the first sample is drawn from a uniform distribution P on the unit area. The other sample is drawn from the perturbed uniform distribution $Q_\omega$ with density $1 + sin(\omega x) sin(\omega y)$. For increasing values of $\omega$, the discrimination of $Q_\omega$ from $P$ becomes harder (Figure 4). The range of $\omega$ changes between 1 to 6. We call these problems Puni1 to Puni6, respectively. The best results for all statistical kernel-based methods are achieved by using a characteristic kernel tailored to the periodic domain, $k(x,y)= \Pi_{i=1}^{2} 1/(1-2\theta cos(x_i-y_i)+\theta ^2)$, with $\theta = 0.9$ tuned using the 5-fold cross validation procedure. The results reported in Table~2 show much smaller values of ARE for RMMD rather than for KFDA and MMD. Figure~5 shows the detailed results of the type~\Rmnum{2} error for RMMD, MMD, and KFDA based on different sample sizes~$n$ and different frequencies~$\omega$. As displayed in Figure~5, RMMD obtains the robust result of zero type~\Rmnum{2} error for 100 samples over all different frequencies. Instead KFDA and MMD need much larger samples for the more difficult cases with larger $\omega$ to obtain a power of one.

\begin{table}[t]
 \caption{The ARE obtained on the periodic data, by applying RMMD with $\kappa=1$, and $\theta = 0.9$ in periodic kernels, and  KFDA with $\gamma=10^{-1}$.}
 \begin{center}
 \begin{normalsize}
 \begin{sc}
 \begin{tabular}{lcccccccccr}
\hline
 & \bf{$e^B_{\MMD,\RMMD}$} & \bf{$e^B_{\MMD,\KFDA}$}  & \bf{$e^B_{\KFDA,\RMMD}$} \\
\hline
Periodic1 & \bf{0.71} & 0.75 & 0.93 \\
Preiodic2 & \bf{0.75} & 1 & \bf{0.75} \\
Puni1 & \bf{0.11} & 0.78 & 0.14 \\
Puni2 & \bf{0.09} & 0.82 & 0.11 \\
Puni3 & \bf{0.09} & 0.82 & 0.11 \\
Puni4 & \bf{0.08} & 0.85 & 0.09 \\
Puni5 & \bf{0.07} & 0.88 & 0.06 \\
Puni6 & \bf{0.05} & 0.81 & 0.06 \\
\hline
\end{tabular}
\end{sc}
\end{normalsize}
\end{center}
\end{table}

\begin{figure*}[!ht]
 \begin{center}
 \centerline{\includegraphics[scale=0.65]{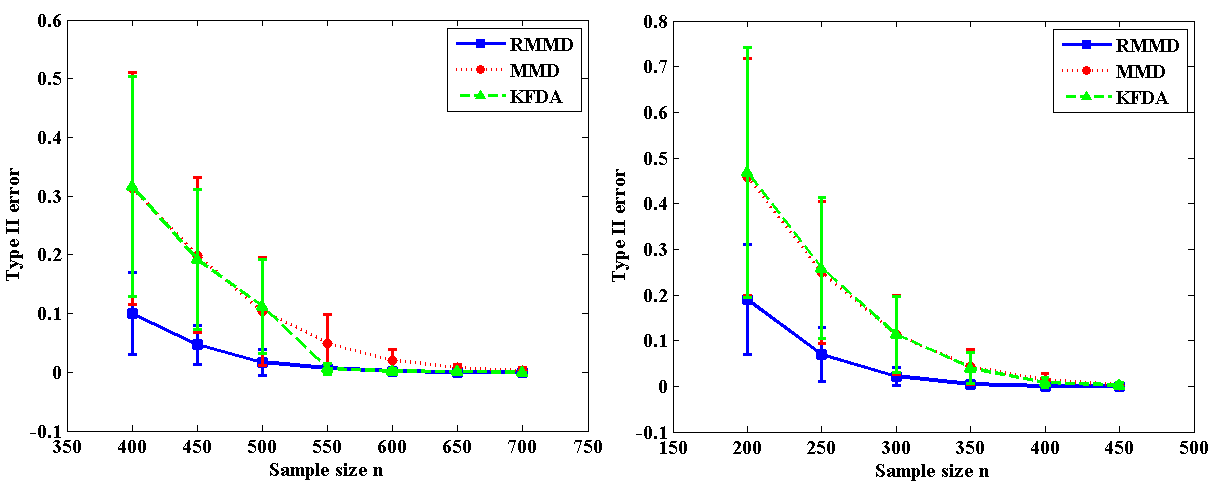}}
 \caption{Type~\Rmnum{2} error change based on different sample size n. On the left, the results with Periodic kernel 1 and on the
right, the results with Periodic kernel 2.}
\label{per1per2}
\end{center}
\end{figure*}
\begin{figure*}[!ht]
 \begin{center}
 \centerline{\includegraphics[scale=0.65]{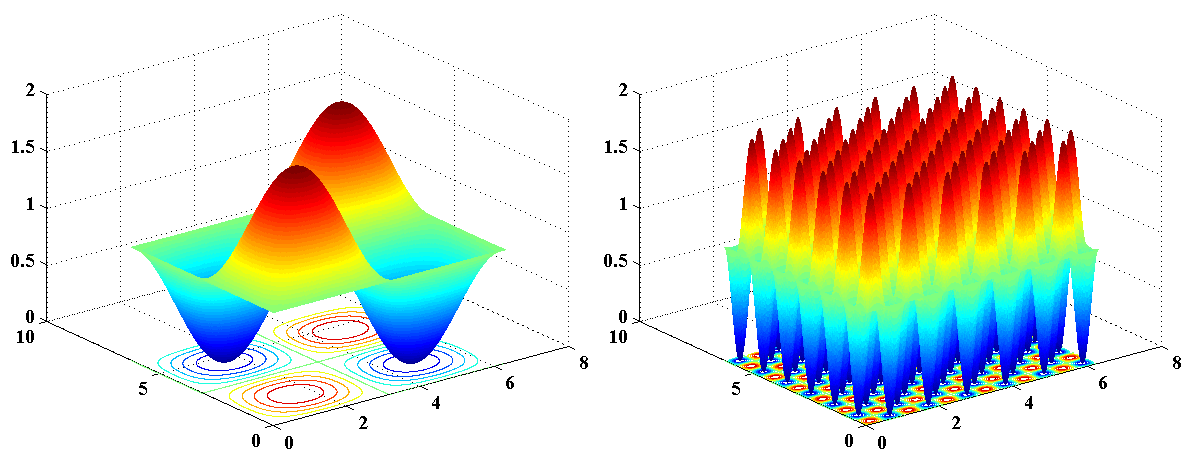}}
 \caption{ The probability density function of Puni1 with $\omega=1$ on the left and the probability density function of Puni6 with $\omega=6$ on the right. As $\omega$ increases the probability density function looks more similar to the uniform distribution and the discrimination of P and $Q_\omega$ becomes more difficult for the test statistics.}
\label{fig3}
\end{center}
\end{figure*}  

\begin{figure*}[!ht]
 \begin{center}
 \centerline{\includegraphics[scale=0.33]{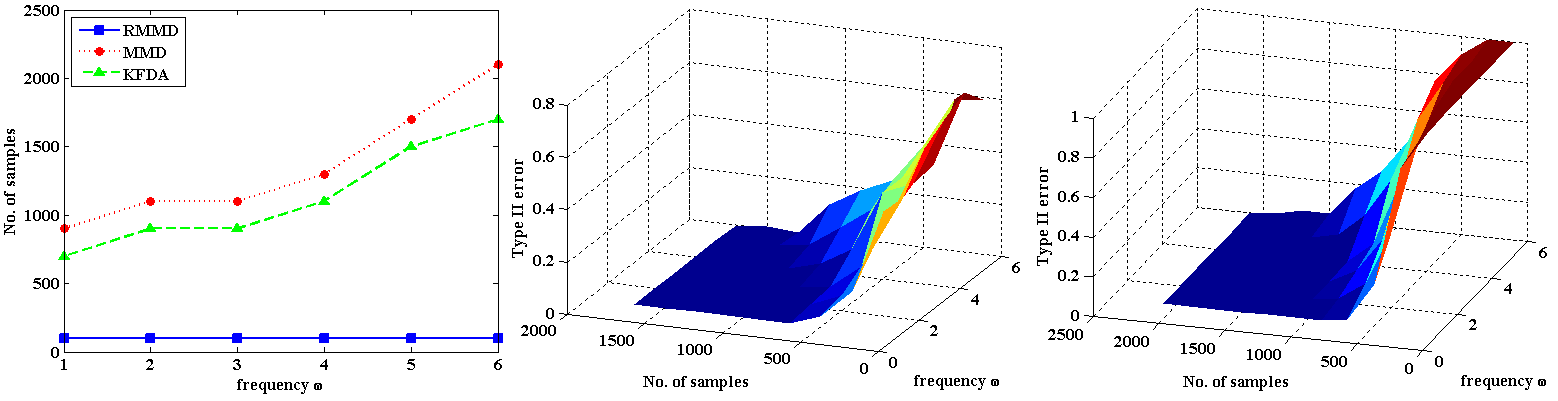}}
 \caption{ On the left, different sample sizes $n$ for different frequencies $\omega$ are shown. The type~\Rmnum{2} error changes based on different sample sizes $n$ and different frequencies $\omega$, in the middle for the KFDA-based test, and on the right for the MMD-based test.}
\label{fig3}
\end{center}
\end{figure*} 
\subsection{MNIST, Covertype, and Flare-Solar Datasets}

Moving from synthetic data to standard benchmarks, we tested our method on three datasets: 1) the MNIST dataset of handwritten digits (LibSVM library: 10 classes, 5000 data points, and 784 dimensions); 2) the Covertype dataset of forest cover types (LibSVM library: 7 classes, 1400 instances, and 54 dimensions); 3) the Flare-Solar dataset (mldata.org: 2 classes, 100 instances, 10 dimensions). We compare the performance of RMMD with $\kappa = 1$, KFDA with $\gamma  = 10^{-1}$ and MMD, using the pairwise approach and testing for differences between the distributions of the classes, see Table~1. We average the results over 10 runs. The family wide level is set to $\alpha = 0.05$ (resulting in $\bar{\alpha} = 0.0011$, $\bar{\alpha} = 0.0024$ and $\bar{\alpha} = 0.05$ for each individual comparison for MNIST, Covertype and Flare-Solar datasets, respectively).
 The RMMD-based test achieves higher power than the other methods (see Table~1).
 
 \subsection{Electroencephalography Data}

We recorded EEG from four subjects performing a visual task. A checkerboard was presented in the subject's left visual field. We refer to \cite{Whittingstall} for details on data collection and preprocessing. In our learning task, for each subject we have 64 signal distributions assigned to 64 electrodes. The data contain 360 instances of a 200 dimensional feature vector for each distribution. The goal of hypothesis testing is to disambiguate signals recorded from electrodes
corresponding to early visual cortex from the rest. This is difficult because of the low signal-to-noise ratio and the similarity of the patterns of all electrodes. Moreover, the high number of electrodes makes this experiment a good candidate to assess the multiple comparison part of our method. In the one-vs-all approach the normalized distribution of each electrode is compared to the normalized combined distribution of the other 63 electrodes. RMMD with $\kappa=1$ with Gaussian kernel is used as our hypothesis test. The parameter $\sigma$ of the Gaussian kernel is set to the median distance of data points. The results of our hypothesis test reject the null hypothesis and confirm the dissimilarity of distributions in 63 electrodes. The results of the pairwise approach with RMMD and MMD are depicted in Figure~6.\\

\begin{figure*}[!ht]
 \begin{center}
 \centerline{\includegraphics[scale=1]{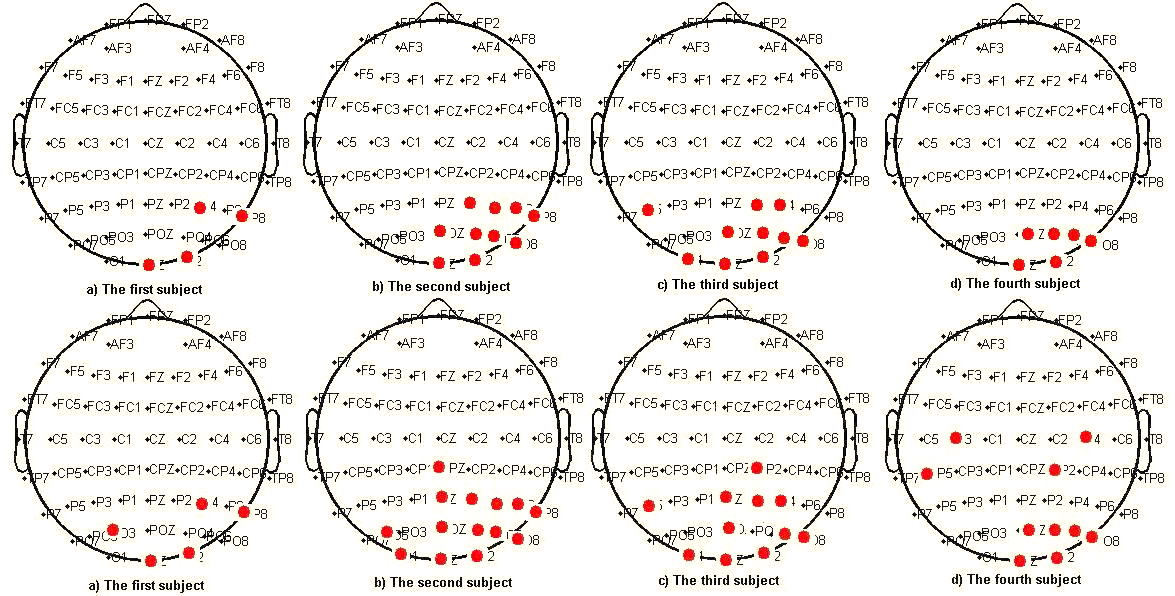}}
 \caption{The results of RMMD and MMD as hypothesis tests on the EEG data recorded from 64 electrodes per subject in the top row and the bottom row, respectively. Categorized electrodes recognized by the two methods as related to the visual task are colored.}
\label{eeg}
\end{center}
\end{figure*} 
\begin{figure*}[!ht]
 \begin{center}
 \centerline{\includegraphics[scale=0.55]{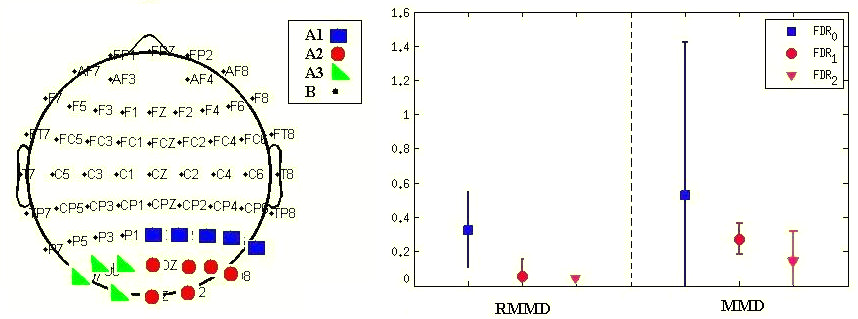}}
 \caption{The reference image of the EEG electrodes is shown on the left. We categorized electrodes into four groups as follows: A1, the electrodes corresponding to visual cortex in the region of interest, A2, the peripheral electrodes that can be wrongly detected due to noise, A3, the electrodes in the left visual cortex often detected due to noise or interrelation between brain areas, and B, all the remaining  electrodes. On the right, the results of RMMD and MMD are quantitatively compared based of the FDRs defined in the text. The smallest and most robust FDRs are obtained by RMMD.}
\label{fig6}
\end{center}
\end{figure*} 

Neuroscientists usually subjectively assess the results obtained from imaging techniques and inferred from machine learning. For instance, in the current experiment the expectation is that electrodes in region A1 are categorized together by means of EEG imaging techniques and multiple comparisons. But electrodes of other area (such as $A_2$ and $A_3$, see Figure~7) can be confused as belonging to $A_1$ due to the high noise. Figure~7 describes the categorization of the electrodes.
We assess our results quantitatively by means of False Discovery Rates (FDR), using the following FDRs to compare the results of RMMD to those of MMD: \\

$FDR_0=
\frac{(\text{no. of electrodes categorized for the visual task in} ~A2\cup A3\cup B)}{U}$, 

$FDR_1=\frac{(\text{no. of electrodes categorized for the visual task in } A3\cup B)}{U},$ 

$FDR_2=\frac{(\text{no. of electrodes categorized for the visual task in } B)}{U}$,\\

where $U$ is the total number of electrodes categorized for the task. The results are depicted in Figure~7. RMMD obtained more robust and better results than MMD with smaller FDRs.

\section{Conclusion}

Our novel regularized maximum mean discrepancy (RMMD) is a kernel-based test statistic generalizing the MMD test. We proved that RMMD overpowers MMD and KFDA; power consistency is obtained with higher rate. Power control makes RMMD a good hypothesis test for multiple comparisons, especially for the crucial case of small sample sizes. In contrast to KFDA and MMD, the convergence of RMMD-based test statistics to the normal distribution under null and alternative hypotheses yields fast and straightforward RMMD estimates. Experiments with goldstandard benchmarks (MNIST, Covertype and Flare-Solar dataset) and with EEG data yield state of the art results.
 
\section{Acknowledgement}
This work was partially funded by the Nano-resolved Multi-scale Investigations of human tactile Sensations and tissue Engineered nanobio-Sensors (SINERGIA:CRSIKO\_122697/1) grant, and the State representation in reward based learning from spiking neuron models to psychophysics(NaNoBio Touch:228844) grant.

\end{document}